%% file: emnlp2020.tex
\pdfoutput=1 %

\documentclass[11pt,a4paper]{article}
\usepackage[hyperref]{emnlp2020}
\usepackage{times}
\usepackage{latexsym}

\usepackage{microtype}

\aclfinalcopy %

\usepackage[utf8]{inputenc}
\usepackage{multirow}
\usepackage{booktabs}
\usepackage{textcomp}  %
\usepackage{adjustbox}
\usepackage{xspace}
\input{tables_nli}

\input{tables_qa}

\newcommand{\roberta}{{\small \textsc{Roberta}}\xspace}
\newcommand{\xlmr}{{\small \textsc{XLM-R}}\xspace}
\newcommand{\orig}{{\small \textsc{Orig}}\xspace}
\newcommand{\ppes}{{\small \textsc{BT-ES}}\xspace}
\newcommand{\ppfi}{{\small \textsc{BT-FI}}\xspace}
\newcommand{\mtes}{{\small \textsc{MT-ES}}\xspace}
\newcommand{\mtfi}{{\small \textsc{MT-FI}}\xspace}
\newcommand{\pp}{{\small \textsc{BT-XX}}\xspace}
\newcommand{\mt}{{\small \textsc{MT-XX}}\xspace}
\newcommand{\translatetest}{{\small \textsc{Translate-Test}}\xspace}
\newcommand{\translatetrain}{{\small \textsc{Translate-Train}}\xspace}

\newcommand{\zeroshot}{{\small \textsc{Zero-Shot}}\xspace}

\title{Translation Artifacts in Cross-lingual Transfer Learning}

\author{Mikel Artetxe, \ Gorka Labaka, \ Eneko Agirre \\
HiTZ Center \\
University of the Basque Country (UPV/EHU) \\
\texttt{\{mikel.artetxe,gorka.labaka,e.agirre\}@ehu.eus}
}

\date{}

\begin{document}
\maketitle
\begin{abstract}
Both human and machine translation play a central role in cross-lingual transfer learning: many multilingual datasets have been created through professional translation services, and using machine translation to translate either the test set or the training set is a widely used transfer technique. In this paper, we show that such translation process can introduce subtle artifacts that have a notable impact in existing cross-lingual models. For instance, in natural language inference, translating the premise and the hypothesis independently can reduce the lexical overlap between them, which current models are highly sensitive to. We show that some previous findings in cross-lingual transfer learning need to be reconsidered in the light of this phenomenon. Based on the gained insights, we also improve the state-of-the-art in XNLI for the \textit{translate-test} and \textit{zero-shot} approaches by 4.3 and 2.8 points, respectively.
\end{abstract}

\section{Introduction}
\label{sec:introduction}

While most NLP resources are English-specific, 
there have been several recent efforts to build \textbf{multilingual benchmarks}.
One possibility is to collect and annotate data in multiple languages separately \citep{clark2020tydiqa}, but most existing datasets have been created through translation \citep{conneau2018xnli,artetxe2019xquad}.
This approach has two desirable properties: it relies on existing professional translation services rather than requiring expertise in multiple languages, and it results in parallel evaluation sets that offer a meaningful measure of the cross-lingual transfer gap of different models. The resulting multilingual datasets are generally used for evaluation only, relying on existing English datasets for training.

Closely related to that,
\textbf{cross-lingual transfer learning} aims to leverage large datasets available in one language---typically English---to build multilingual models that can generalize to other languages.
Previous work has explored 3 main approaches to that end: machine translating the test set into English and using a monolingual English model (\translatetest), machine translating the training set into each target language and training the models on their respective languages (\translatetrain),
or using English data to fine-tune a multilingual model that is then transferred to the rest of languages (\zeroshot).

The dataset creation and transfer procedures described above result in a \textbf{mixture of original,\footnote{We use the term \textit{original} to refer to non-translated text.} human translated and machine translated data} when dealing with cross-lingual models. In fact, the type of text a system is trained on does not typically match the type of text it is exposed to at test time: \translatetest systems are trained on original data and evaluated on machine translated test sets, \zeroshot systems are trained on original data and evaluated on human translated test sets, and \translatetrain systems are trained on machine translated data and evaluated on human translated test sets.

Despite overlooked to date, we show that \textbf{such mismatch has a notable impact} in the performance of existing cross-lingual models. By using back-translation \citep{sennrich2016improving} to paraphrase each training instance, we obtain another English version of the training set
that better resembles the test set, obtaining substantial improvements for the \translatetest and \zeroshot approaches in cross-lingual Natural Language Inference (NLI). While improvements brought by machine translation have previously been attributed to data augmentation \citep{singh2019xlda}, we reject this hypothesis and show that the phenomenon is only present in translated test sets, but not in original ones.
Instead, our analysis reveals that this behavior is caused by subtle \textbf{artifacts arising from the translation} process itself. In particular, we show that translating different parts of each instance separately (e.g., the premise and the hypothesis in NLI) can alter superficial patterns in the data (e.g., the degree of lexical overlap between them), which severely affects the generalization ability of current models.
Based on the gained insights, we improve the state-of-the-art in XNLI, and show that some previous findings need to be reconsidered in the light of this phenomenon.

\section{Related work}
\label{sec:related_work}

\paragraph{Cross-lingual transfer learning.}
Current cross-lingual models work by
pre-training multilingual representations using some form of language modeling, which are then fine-tuned on the relevant task and transferred to different languages.
Some authors leverage parallel data to that end \citep{conneau2019xlm,huang2019unicoder}, but
training a model akin to BERT \citep{devlin2019bert} on the combination of monolingual corpora in multiple languages is also effective \citep{conneau2019xlmr}. Closely related to our work, \citet{singh2019xlda} showed that replacing segments of the training data with their translation during fine-tuning is helpful.
However, they attribute this behavior to a data augmentation effect, which we believe should be reconsidered given the new evidence we provide. %

\paragraph{Multilingual benchmarks.} Most benchmarks covering a wide set of languages have been created through translation, as it is the case of XNLI \citep{conneau2018xnli} for NLI, PAWS-X \citep{yang2019pawsx} for adversarial paraphrase identification, and XQuAD \citep{artetxe2019xquad} and MLQA \citep{lewis2019mlqa} for Question Answering (QA). A notable exception is TyDi QA \citep{clark2020tydiqa}, a contemporaneous QA dataset that was separately annotated in 11 languages.
Other cross-lingual datasets leverage existing multilingual resources, as it is the case of MLDoc \citep{schwenk2018corpus} for document classification and Wikiann \citep{pan2017wikiann} for named entity recognition.
Concurrent to our work, \citet{hu2020xtreme} combine some of these datasets into a single multilingual benchmark, and evaluate some well-known methods on it.

\paragraph{Annotation artifacts.} Several studies have shown that NLI datasets like SNLI \citep{bowman2015snli} and MultiNLI  \citep{williams2018broad} contain spurious patterns that can be exploited to obtain strong results without making real inferential decisions. For instance, \citet{gururangan2018annotation} and \citet{poliak2018hypothesis} showed that a hypothesis-only baseline performs better than chance due to cues on their lexical choice and sentence length.
Similarly, \citet{mccoy2019right} showed that NLI models tend to predict \textit{entailment} for sentence pairs with a high lexical overlap.
Several authors have worked on adversarial datasets to diagnose these issues and provide a more challenging benchmark \citep{naik2018stress,glockner2018breaking,nie2019adversarial}. Besides NLI,
other tasks like QA have also been found to be susceptible to annotation artifacts \citep{jia2017adversarial,kaushik2018reading}.
While previous work has focused on the monolingual scenario, we show that translation can interfere with these artifacts in multilingual settings.

\paragraph{Translationese.} Translated texts are known to have unique features like simplification, explicitation, normalization and interference, which are refer to as \textit{translationese} \citep{volansky2013translationese}.
This phenomenon has been reported to have a notable impact in machine translation evaluation \citep{zhang2019effect,graham2019translationese}.
For instance, back-translation brings large BLEU gains for reversed test sets (i.e., when translationese is on the source side and original text is used as reference), but its effect diminishes in the natural direction \citep{edunov2019evaluation}.
While connected, the phenomenon we analyze is different in that it arises from translation inconsistencies due to the lack of context, and affects cross-lingual transfer learning rather than machine translation.

\section{Experimental design}
\label{sec:design}

Our goal is to analyze the effect of both human and machine translation in cross-lingual models. For that purpose, the core idea of our work is to (i) use machine translation to either translate the training set into other languages, or generate English paraphrases of it through back-translation, and
(ii) evaluate the resulting systems on original, human translated and machine translated test sets in comparison with systems trained on original data.
We next describe the models used in our experiments (\S \ref{subsec:models}), the specific training variants explored (\S \ref{subsec:variants}), and the evaluation procedure followed (\S \ref{subsec:tasks}).

\subsection{Models and transfer methods} \label{subsec:models}

We experiment with two models that are representative of the state-of-the-art in monolingual and cross-lingual pre-training: (i) \roberta \citep{liu2019roberta}, which is an improved version of BERT that uses masked language modeling to pre-train an English Transformer model, and (ii) \xlmr \citep{conneau2019xlmr}, which is a multilingual extension of the former pre-trained on 100 languages.
In both cases, we use the large models released by the authors under the fairseq repository.\footnote{\url{https://github.com/pytorch/fairseq}} As discussed next, we explore different variants of the training set to fine-tune each model on different tasks. At test time, we try both machine translating the test set into English (\translatetest) and, in the case of \xlmr, using the actual test set in the target language (\zeroshot).

\subsection{Training variants} \label{subsec:variants}

We try 3 variants of each training set to fine-tune our models: (i) the original one in English (\orig), (ii) an English paraphrase of it generated through back-translation using Spanish or Finnish as pivot (\ppes and \ppfi), and (iii) a machine translated version in Spanish or Finnish (\mtes and \mtfi).
For sentences occurring multiple times in the training set (e.g., premises repeated for multiple hypotheses), we use the exact same translation for all occurrences, as our goal is to understand the inherent effect of translation rather than its potential application as a data augmentation method.

In order to train the machine translation systems for \mt and \pp, we use the big Transformer model \citep{vaswani2017attention} with the same settings as \citet{ott2018scaling} and SentencePiece tokenization \citep{kudo2018sentencepiece} with a joint vocabulary of 32k subwords. For English-Spanish, we train for 10 epochs on all parallel data from WMT 2013 \citep{bojar2013findings} and ParaCrawl v5.0 \citep{espla2019paracrawl}. For English-Finnish, we train for 40 epochs on Europarl and Wiki Titles from WMT 2019 \citep{barrault2019findings}, ParaCrawl v5.0, and DGT, EUbookshop and TildeMODEL from OPUS \citep{tiedemann2012opus}. In both cases, we remove sentences longer than 250 tokens, with a source/target ratio exceeding 1.5, or for which \texttt{langid.py} \citep{lui2012langid} predicts a different language, resulting in a final corpus size of 48M and 7M sentence pairs, respectively. We use sampling decoding with a temperature of 0.5 for inference, which produces more diverse translations than beam search \citep{edunov2018understanding} and performed better in our preliminary experiments.

\begin{table*}[ht!]
\begin{center}
\addtolength{\tabcolsep}{-1.5pt}
\resizebox{1\linewidth}{!}{
\insertXnliDevTable
}
\caption{\textbf{XNLI dev results (acc).} \pp and \mt consistently outperform \orig in all cases.
\label{tab:xnli_dev}}
\end{center}
\end{table*}

\subsection{Tasks and evaluation procedure} \label{subsec:tasks}

We use the following tasks for our experiments:

\paragraph{Natural Language Inference (NLI).} Given a premise and a hypothesis, the task is to determine whether there is an \textit{entailment}, \textit{neutral} or \textit{contradiction} relation between them. We fine-tune our models on MultiNLI \citep{williams2018broad} for 10 epochs using the same settings as \citet{liu2019roberta}. In most of our experiments, we evaluate on XNLI \citep{conneau2018xnli}, which comprises 2490 development and 5010 test instances in 15 languages. These were originally annotated in English,
and the resulting premises and hypotheses were independently translated into the rest of the languages by professional translators. For the \translatetest approach, we use the machine translated versions from the authors. Following \citet{conneau2019xlmr}, we select the best epoch checkpoint according to the average accuracy in the development set.  %

\paragraph{Question Answering (QA).} Given a context paragraph and a question, the task is to identify the span answering the question in the context. We fine-tune our models on SQuAD v1.1 \citep{rajpurkar2016squad} for 2 epochs using the same settings as \citet{liu2019roberta}, and report test results for the last epoch.
We use two datasets for evaluation: XQuAD \citep{artetxe2019xquad}, a subset of the SQuAD development set translated into 10 other languages, and MLQA \citep{lewis2019mlqa} a dataset consisting of parallel context paragraphs plus the corresponding questions annotated in English and translated into 6 other languages. In both cases, the translation was done by professional translators at the document level (i.e., when translating a question, the text answering it was also shown).
For our \pp and \mt variants, we translate the context paragraph and the questions independently, and map the answer spans using the same procedure as \citet{carrino2019automatic}.\footnote{We use FastAlign \citep{dyer2013simple} for word alignment, and discard the few questions for which the mapping method fails (when none of the tokens in the answer span are aligned).}
For the \translatetest approach, we use the official machine translated versions of MLQA, run inference over them, and map the predicted answer spans back to the target language.\footnote{We use the same procedure as for the training set except that (i) given the small size of the test set, we combine it with WikiMatrix \citep{schwenk2019wikimatrix} to aid word alignment, (ii) we use Jieba for Chinese segmentation instead of the Moses tokenizer, and (iii) for the few unaligned spans, we return the English answer.}

Both for NLI and QA, we run each system 5 times with different random seeds and report the average results. Space permitting, we also report the standard deviation across the 5 runs. In our result tables, we use an underline to highlight the best result within each block, and boldface to highlight the best overall result.

\section{NLI experiments}
\label{sec:nli}

We next discuss our main results in the XNLI development set (\S \ref{subsec:translatetest}, \S \ref{subsec:zeroshot}), run additional experiments to better understand the behavior of our different variants (\S \ref{subsec:translationese}, \S \ref{subsec:stress}, \S \ref{subsec:distribution}), and compare our results to previous work in the XNLI test set (\S \ref{subsec:sota}).

\subsection{\translatetest results} \label{subsec:translatetest}

We start by analyzing XNLI development results for \translatetest. Recall that, in this approach, the test set is machine translated into English, but training is typically done on original English data. Our \ppes and \ppfi variants close this gap by training on a machine translated English version of the training set generated through back-translation. As shown in Table \ref{tab:xnli_dev}, this brings substantial gains for both \roberta and \xlmr, with an average improvement of 4.6 points in the best case. %
Quite remarkably, \mtes and \mtfi also outperform \orig by a substantial margin, and are only 0.8 points below their \ppes and \ppfi counterparts. Recall that, for these two systems, training is done in machine translated Spanish or Finnish, while inference is done in machine translated English. This shows that the loss of performance when generalizing from original data to machine translated data is substantially larger than the loss of performance when generalizing from one language to another.

\subsection{\zeroshot results}  \label{subsec:zeroshot}

We next analyze the results for the \zeroshot approach. In this case, inference is done in the test set in each target language which, in the case of XNLI, was human translated from English. As such, different from the \translatetest approach, neither training on original data (\orig) nor training on machine translated data (\pp and \mt) makes use of the exact same type of text that the system is exposed to at test time.
However, as shown in Table \ref{tab:xnli_dev}, both \pp and \mt outperform \orig by approximately 2 points, which suggests that our (back-)translated versions of the training set are more similar to the human translated test sets than the original one. %
This also provides a new perspective on the \translatetrain approach, which was reported to outperform \orig in previous work \citep{conneau2019xlm}: while the original motivation was to train the model on the same language that it is tested on, our results show that machine translating the training set is beneficial even when the target language is different.

\subsection{Original vs. translated test sets}  \label{subsec:translationese}

\begin{table}
\begin{center}
\addtolength{\tabcolsep}{-1.5pt}
\resizebox{0.84\linewidth}{!}{
\insertTranslateTable
}
\end{center}
\caption{\textbf{NLI results on original (OR), human translated (HT) and machine translated (MT) sets (acc).} \pp and \mt outperform \orig in translated sets, but do not get any clear improvement in original ones.
\label{tab:translationese}}
\end{table}

So as to understand whether the improvements observed so far are limited to translated test sets or apply more generally, we conduct additional experiments comparing translated test sets to original ones.
However, to the best of our knowledge, all existing non-English NLI benchmarks were created through translation. For that reason, we build a new test set that mimics XNLI, but is annotated in Spanish rather than English. We first collect the premises from a filtered version of CommonCrawl \citep{buck2014ngram}, taking a subset of 5 websites that represent a diverse set of genres: a newspaper, an economy forum, a celebrity magazine, a literature blog, and a consumer magazine. We then ask native Spanish annotators to generate an \textit{entailment}, a \textit{neutral} and a \textit{contradiction} hypothesis for each premise.\footnote{Unlike XNLI, we do not collect 4 additional labels for each example. Note, however, that XNLI kept the original label as the gold standard, so the additional labels are irrelevant for the actual evaluation. This is not entirely clear in \citet{conneau2018xnli}, but can be verified by inspecting the dataset.} We collect a total of 2490 examples using this procedure, which is the same size as the XNLI development set. Finally, we create
a human translated and a machine translated English version of the dataset using professional translators from Gengo and our machine translation system described in \S \ref{subsec:variants},\footnote{We use beam search instead of sampling decoding.} respectively.
We report results for the best epoch checkpoint on each set.

As shown in Table \ref{tab:translationese}, both \pp and \mt clearly outperform \orig in all test sets created through translation, which is consistent with our previous results. In contrast, the best results on the original English set are obtained by \orig, and neither \pp nor \mt obtain any clear improvement on the one in Spanish either.\footnote{Note that the standard deviations are around 0.3.}
This confirms that the underlying phenomenon is limited to translated test sets.
In addition, it is worth mentioning that the results for the machine translated test set in English are slightly better than those for the human translated one,
which suggests that the difficulty of the task does not only depend on the translation quality.
Finally, it is also interesting that \mtes is only marginally better than \mtfi in both Spanish test sets, even if it corresponds to the \translatetrain approach, whereas \mtfi needs to \zeroshot transfer from Finnish into Spanish.
This reinforces the idea that it is training on translated data rather than training on the target language that is key in \translatetrain.

\subsection{Stress tests}  \label{subsec:stress}

\begin{table}
\begin{center}
\addtolength{\tabcolsep}{-1.5pt}
\resizebox{1\linewidth}{!}{
\insertStressTestTable
}
\end{center}
\caption{\textbf{NLI Stress Test results (combined matched \& mismatched acc).} AT = antonymy, NR = numerical reasoning, WO = word overlap, NG = negation, LN = length mismatch, SE = spelling error. \ppfi and \mtfi are considerably weaker than \orig in the competence test, but substantially stronger in the distraction test.
\label{tab:stress}}
\end{table}

In order to better understand how systems trained on original and translated data differ, we run additional experiments on the NLI Stress Tests \citep{naik2018stress}, which were designed to test the robustness of NLI models to specific linguistic phenomena in English. The benchmark consists of a competence test, which evaluates the ability to understand antonymy relation and perform numerical reasoning, a distraction test, which evaluates the robustness to shallow patterns like lexical overlap and the presence of negation words, and a noise test, which evaluates robustness to spelling errors. Just as with previous experiments, we report results for the best epoch checkpoint in each test set.

As shown in Table \ref{tab:stress}, \orig outperforms \ppfi and \mtfi on the competence test by a large margin, but the opposite is true on the distraction test.\footnote{We observe similar trends for \ppes and \mtes, but omit these results for conciseness.}
In particular, our results show that \ppfi and \mtfi are less reliant on lexical overlap and the presence of negative words. This feels intuitive, as translating the premise and hypothesis independently---as \ppfi and \mtfi do---is likely to reduce the lexical overlap between them.
More generally, the translation process can alter similar superficial patterns in the data, which NLI models are sensitive to (\S \ref{sec:related_work}). This would explain why the resulting models have a different behavior on different stress tests.

\subsection{Output class distribution}  \label{subsec:distribution}

With the aim to understand the effect of the previous phenomenon in cross-lingual settings, we look at the output class distribution of our different models in the XNLI development set.
As shown in Table \ref{tab:distribution}, the predictions of all systems are close to the true class distribution in the case of English. Nevertheless, \orig is strongly biased for the rest of languages, and tends to underpredict \textit{entailment} and overpredict \textit{neutral}. This can again be attributed to the fact that the English test set is original, whereas the rest are human translated. In particular, it is well-known that NLI models tend to predict \textit{entailment} when there is a high lexical overlap between the premise and the hypothesis (\S \ref{sec:related_work}).
However, the degree of overlap will be smaller in the human translated test sets given that the premise and the hypothesis were translated independently, which explains why \textit{entailment} is underpredicted.
In contrast, \ppfi and \mtfi are exposed to the exact same phenomenon during training, which explains why they are not that heavily affected.

So as to measure the impact of this phenomenon, we explore a simple approach to correct this bias: having fine-tuned each model, we adjust the bias term added to the logit of each class so the model predictions match the true class distribution for each language.\footnote{We achieve this using an iterative procedure where, at each step, we select one class and set its bias term so the class is selected for the right percentage of examples.} As shown in Table \ref{tab:bias}, this brings large improvements for \orig, but is less effective for \ppfi and \mtfi.\footnote{Note that we are adjusting the bias term in the evaluation set itself, which requires knowing its class distribution and is thus a form of cheating. While useful for analysis, a fair comparison would require adjusting the bias term in a separate validation set. This is what we do for our final results in \S \ref{subsec:sota}, where we adjust the bias term in the XNLI development set and report results on the XNLI test set.} This shows that the performance of \orig was considerably hindered by this bias, which \ppfi and \mtfi effectively mitigate.

\begin{table}
\begin{center}
\addtolength{\tabcolsep}{-2.5pt}
\resizebox{1\linewidth}{!}{
\insertClassDistributionTable
}
\end{center}
\caption{\textbf{Output class distribution on XNLI dev.} All systems are close to the true distribution in English, but \orig is biased toward \textit{neu} and \textit{con} in the transfer languages. \ppfi and \mtfi alleviate this issue.
\label{tab:distribution}}
\end{table}

\begin{table}
\begin{center}
\addtolength{\tabcolsep}{-1.5pt}
\resizebox{0.83\linewidth}{!}{
\insertBiasTable
}
\end{center}
\caption{\textbf{XNLI dev results with class distribution unbiasing (average acc across all languages).} Adjusting the bias term of the classifier to match the true class distribution brings large improvements for \orig, but is less effective for \ppfi and \mtfi.
\label{tab:bias}}
\end{table}

\subsection{Comparison with the state-of-the-art}  \label{subsec:sota}

\begin{table*}
\begin{center}
\addtolength{\tabcolsep}{-1.5pt}
\resizebox{1\linewidth}{!}{
\insertXnliSotaTable
}
\caption{\textbf{XNLI test results (acc).}
Results for other methods are taken from their respective papers or, if not provided, from \citet{conneau2019xlmr}. For those with multiple variants, we select the one with the best results.
\label{tab:xnli_test}}
\end{center}
\end{table*}

So as to put our results into perspective, we compare our best variant to previous work on the XNLI test set. As shown in Table \ref{tab:xnli_test}, our method improves the state-of-the-art for both the \translatetest and the \zeroshot approaches by 4.3 and 2.8 points, respectively. It also obtains the best overall results published to date, with the additional advantage that the previous state-of-the-art required a machine translation system between English and each of the 14 target languages, whereas our method uses a single machine translation system between English and Finnish (which is not one of the target languages). While the main goal of our work is not to design better cross-lingual models, but to analyze their behavior in connection to translation, this shows that the phenomenon under study is highly relevant, to the extent that it can be exploited to improve the state-of-the-art.

\section{QA experiments}
\label{sec:qa}

So as to understand whether our previous findings apply to other tasks besides NLI, we run additional experiments on QA. As shown in Table \ref{tab:mlqa}, \ppfi and \ppes do indeed outperform \orig for the \translatetest approach on MLQA. The improvement is modest, but very consistent across different languages, models and runs. The results for \mtes and \mtfi are less conclusive, presumably because mapping the answer spans across languages might introduce some noise. In contrast, we do not observe any clear improvement for the \zeroshot approach on this dataset. Our XQuAD results in Table \ref{tab:xquad_f1} are more positive, but still inconclusive.

These results can partly be explained by the translation procedure used to create the different benchmarks: the premises and hypotheses of XNLI were translated independently, whereas the questions and context paragraphs of XQuAD were translated together. Similarly, MLQA made use of parallel contexts, and translators were shown the sentence containing each answer when translating the corresponding question. As a result, one can expect both QA benchmarks to have more consistent translations than XNLI, which would in turn diminish this phenomenon. In contrast, the questions and context paragraphs are independently translated when using machine translation, which explains why \ppes and \ppfi outperform \orig for the \translatetest approach. %
We conclude that the translation artifacts revealed by our analysis are not exclusive to NLI, as they also show up on QA for the \translatetest approach, but their actual impact can be highly dependent on the translation procedure used and the nature of the task.

\begin{table*}
\begin{center}
\addtolength{\tabcolsep}{-1.5pt}
\resizebox{0.93\linewidth}{!}{
\insertMlqaTable
}
\end{center}
\caption{\textbf{MLQA test results (F1 / exact match).}
\label{tab:mlqa}}
\end{table*}

\begin{table*}
\begin{center}
\addtolength{\tabcolsep}{-1.5pt}
\resizebox{0.72\linewidth}{!}{
\insertXquadFTable
}
\end{center}
\caption{\textbf{XQuAD results (F1).} Results for the exact match metric are similar.
\label{tab:xquad_f1}}
\end{table*}

\section{Discussion}
\label{sec:discussion}

Our analysis prompts to reconsider previous findings in cross-lingual transfer learning as follows:

\paragraph{The cross-lingual transfer gap on XNLI was overestimated.} Given the parallel nature of XNLI,
accuracy differences across languages are commonly interpreted as the loss of performance when generalizing from English to the rest of languages.
However, our work shows that there is another factor that can have a much larger impact: the loss of performance when generalizing from original to translated data. Our results suggest that the real cross-lingual generalization ability of \xlmr is considerably better than what the accuracy numbers in XNLI reflect.

\paragraph{Overcoming the cross-lingual gap is not what makes \translatetrain work.} The original motivation for \translatetrain was to train the model on the same language it is tested on. However, we show that it is training on translated data, rather than training on the target language, that is key for this approach to outperform \zeroshot as reported by previous authors.

\paragraph{Improvements previously attributed to data augmentation should be reconsidered.} The method by \citet{singh2019xlda} combines machine translated premises and hypotheses in different languages (\S \ref{sec:related_work}), resulting in an effect similar to \pp and \mt.
As such, we believe that this method should be analyzed from the point of view of dataset artifacts rather than data augmentation, as the authors do.\footnote{Recall that our experimental design prevents a data augmentation effect, in that the number of unique sentences and examples used for training is always the same (\S \ref{subsec:variants}).}
From this perspective, having the premise and the hypotheses in different languages can reduce the superficial patterns between them, which would explain why this approach is better than using examples in a single language.

\paragraph{The potential of \translatetest was underestimated.}
The previous best results for \translatetest on XNLI lagged behind the state-of-the-art by 4.6 points. Our work reduces this gap to only 0.8 points by addressing the underlying translation artifacts. The reason why \translatetest is more severely affected by this phenomenon is twofold: (i) the effect is doubled by first using human translation to create the test set and then machine translation to translate it back to English, and (ii) \translatetrain was inadvertently mitigating this issue (see above), but equivalent techniques were never applied to \translatetest.

\paragraph{Future evaluation should better account for translation artifacts.} The evaluation issues raised by our analysis do not have a simple solution. In fact, while we use the term \textit{translation artifacts} to highlight that they are an unintended effect of translation that impacts final evaluation, one could also argue that it is the original datasets that contain the artifacts, which translation simply alters or even mitigates.\footnote{For instance, the high lexical overlap observed for the \textit{entailment} class is usually regarded a spurious pattern, so reducing it could be considered a positive effect of translation.} In any case, this is a more general issue that falls beyond the scope of cross-lingual transfer learning, so we argue that it should be carefully controlled when evaluating cross-lingual models. In the absence of more robust datasets, we recommend that future multilingual benchmarks should at least provide consistent test sets for English and the rest of languages. This can be achieved by (i) using original annotations in all languages, (ii) using original annotations in a non-English language and translating them into English and other languages, or (iii) if translating from English, doing so at the document level to minimize translation inconsistencies.

\section{Conclusions}
\label{sec:conclusions}

In this paper, we have shown that both human and machine translation can alter superficial patterns in data, which requires reconsidering previous findings in cross-lingual transfer learning. Based on the gained insights, we have improved the state-of-the-art in XNLI for the \translatetest and \zeroshot approaches by a substantial margin.
Finally, we have shown that the phenomenon is not specific to NLI but also affects QA, although
it is less pronounced there thanks to the translation procedure used in the corresponding benchmarks.
So as to facilitate similar studies in the future, we release our NLI dataset,\footnote{\url{https://github.com/artetxem/esxnli}} which, unlike previous benchmarks, was annotated in a non-English language and human translated into English.

\section*{Acknowledgments}

We thank Nora Aranberri and Uxoa Iñurrieta for helpful discussion during the development of this work, as well as the rest of our colleagues from the IXA group that worked as annotators for our NLI dataset.

This research was partially funded by a Facebook Fellowship, the Basque Government excellence research group (IT1343-19), the Spanish MINECO (UnsupMT TIN2017‐91692‐EXP MCIU/AEI/FEDER, UE), Project BigKnowledge (Ayudas Fundación BBVA a equipos de investigación científica 2018), and the NVIDIA GPU grant program.

This research is supported via the IARPA BETTER Program contract No. 2019-19051600006 (ODNI, IARPA activity). The views and conclusions contained herein are those of the authors and should not be interpreted as necessarily representing the official policies, either expressed or implied, of ODNI, IARPA, or the U.S. Government. The U.S. Government is authorized to reproduce and distribute reprints for governmental purposes notwithstanding any copyright annotation therein.

\bibliography{emnlp2020}
\bibliographystyle{acl_natbib}

\end{document}

%% file: tables_nli.tex
\newcommand{\insertXnliDevTable}{
\begin{tabular}[b]{cl|ccccccccccccccc|c}
\toprule
Model & Train & en & fr & es & de & el & bg & ru & tr & ar & vi & th & zh & hi & sw & ur & avg \\
\midrule
\multicolumn{2}{c}{} & \multicolumn{15}{c}{\it Test set machine translated into English (TRANSLATE-TEST)} & \\
\midrule
\multirow{3}{*}{\shortstack{\roberta}}
& \orig & 91.2 & 82.2 & 84.6 & 82.4 & 82.1 & 82.1 & 79.2 & 76.5 & 77.4 & 73.8 & 73.4 & 76.7 & 70.5 & 67.2 & 66.8 & 77.7\scriptsize{ \textpm 0.6} \\
& \ppes & \textbf{\underline{91.6}} & 85.7 & \textbf{\underline{87.4}} & 85.4 & 85.1 & 85.1 & 83.6 & 81.3 & 81.5 & 78.7 & 78.2 & 81.1 & 76.3 & 72.7 & 71.5 & 81.7\scriptsize{ \textpm 0.2} \\
& \ppfi & 91.4 & \textbf{\underline{86.0}} & \textbf{\underline{87.4}} & \underline{85.7} & \textbf{\underline{85.7}} & \underline{85.4} & \textbf{\underline{84.4}} & \underline{82.3} & \underline{82.1} & \underline{79.0} & \underline{79.3} & \underline{81.8} & \underline{77.6} & \underline{73.5} & \underline{73.6} & \underline{82.3}\scriptsize{ \textpm 0.2} \\
\midrule
\multirow{5}{*}{\shortstack{\xlmr}}
& \orig & \underline{90.3} & 82.2 & 84.2 & 82.6 & 81.9 & 82.0 & 79.3 & 76.7 & 77.5 & 75.0 & 73.7 & 77.5 & 70.9 & 67.8 & 67.2 & 77.9\scriptsize{ \textpm 0.3} \\
& \ppes & 90.2 & 84.1 & \underline{86.3} & 84.5 & \underline{84.5} & 84.1 & 82.2 & 79.6 & 80.7 & 78.5 & 77.3 & 80.8 & 75.2 & 72.5 & 71.2 & 80.8\scriptsize{ \textpm 0.3} \\
& \ppfi & 89.5 & \underline{84.9} & 85.5 & 84.5 & \underline{84.5} & \underline{84.6} & \underline{82.9} & \underline{80.6} & \underline{81.4} & \underline{78.9} & \underline{78.1} & \underline{81.5} & \underline{76.3} & \underline{73.3} & \underline{72.5} & \underline{81.3}\scriptsize{ \textpm 0.2} \\
& \mtes & 89.8 & 83.2 & 85.6 & 84.2 & 84.0 & 83.6 & 81.6 & 78.4 & 79.3 & 77.6 & 76.7 & 80.0 & 74.3 & 71.3 & 70.1 & 80.0\scriptsize{ \textpm 0.6} \\
& \mtfi & 89.8 & 84.4 & 85.3 & \underline{84.7} & 84.1 & 84.0 & 82.0 & 79.8 & 80.3 & 77.4 & 77.7 & 80.6 & 74.7 & 71.8 & 71.3 & 80.5\scriptsize{ \textpm 0.3} \\
\midrule
\multicolumn{2}{c}{} & \multicolumn{15}{c}{\it Test set in target language (ZERO-SHOT)} & \\
\midrule
\multirow{5}{*}{\shortstack{\xlmr}}
& \orig & \underline{90.4} & 84.4 & 85.5 & 84.3 & 81.9 & 83.6 & 80.1 & 80.1 & 79.8 & 81.8 & 78.3 & 80.3 & 77.7 & 72.8 & 74.5 & 81.0\scriptsize{ \textpm 0.2} \\
& \ppes & 90.2 & \textbf{\underline{86.0}} & 86.9 & \textbf{\underline{86.5}} & 84.0 & 85.3 & 83.2 & 82.5 & 82.7 & 83.7 & 80.7 & 83.0 & 79.7 & 75.6 & 77.1 & 83.1\scriptsize{ \textpm 0.2} \\
& \ppfi & 89.5 & \textbf{\underline{86.0}} & 86.2 & 86.2 & 83.9 & 85.1 & \underline{83.4} & 82.2 & \textbf{\underline{83.0}} & \textbf{\underline{83.9}} & \textbf{\underline{81.2}} & \textbf{\underline{83.9}} & 80.1 & 75.2 & \textbf{\underline{78.1}} & 83.2\scriptsize{ \textpm 0.1} \\
& \mtes & 89.9 & 85.7 & \underline{87.3} & 85.6 & 83.9 & 85.4 & 82.9 & 82.0 & 82.3 & 83.6 & 80.0 & 82.6 & 79.9 & 75.5 & 76.8 & 82.9\scriptsize{ \textpm 0.4} \\
& \mtfi & 90.2 & 85.9 & 86.9 & \textbf{\underline{86.5}} & \underline{84.4} & \textbf{\underline{85.5}} & \underline{83.4} & \textbf{\underline{83.0}} & 82.4 & 83.6 & 80.5 & 83.6 & \textbf{\underline{80.4}} & \textbf{\underline{76.5}} & 77.9 &  \textbf{\underline{83.4}}\scriptsize{ \textpm 0.2} \\
\bottomrule
\end{tabular}
}

\newcommand{\insertTranslateTable}{
\begin{tabular}[b]{clcccccc}
\toprule
&& \multicolumn{2}{c}{XNLI dev} && \multicolumn{3}{c}{Our dataset} \\
\cmidrule{3-4} \cmidrule{6-8}
&& OR & HT && OR & HT & MT \\
Model & Train & (en) & (es) && (es) & (en) & (en) \\
\midrule
\multirow{3}{*}{\shortstack{\roberta}}
& \orig & \textbf{\underline{92.1}} & - &  & - & 78.7 & 79.0 \\
& \ppes & 91.9 & - &  & - & 80.3 & \textbf{\underline{80.5}} \\
& \ppfi & 91.4 & - &  & - & \textbf{\underline{80.5}} & \textbf{\underline{80.5}} \\
\midrule
\multirow{5}{*}{\shortstack{\xlmr}}
& \orig & \underline{90.5} & 85.5 &  & 81.0 & 77.5 & 78.5 \\
& \ppes & 90.3 & 87.1 &  & \textbf{\underline{81.4}} & 78.6 & \underline{79.4} \\
& \ppfi & 89.7 & 86.5 &  & 80.8 & \underline{78.8} & 79.2 \\
& \mtes & 90.2 & \textbf{\underline{87.5}} &  & 81.3 & 78.4 & 78.9 \\
& \mtfi & 90.4 & 87.1 &  & 81.1 & 78.3 & 78.9 \\
\bottomrule
\end{tabular}
}

\newcommand{\insertStressTestTable}{
\begin{tabular}[b]{clrccrcccrc}
\toprule
&&& \multicolumn{2}{c}{Competence} && \multicolumn{3}{c}{Distraction} && \multicolumn{1}{c}{Noise} \\
\cmidrule{3-3} \cmidrule{4-5} \cmidrule{7-9} \cmidrule{11-11}
Model & Train && AT & NR && WO & NG & LN && SE \\
\midrule
\multirow{2}{*}{\shortstack{\roberta}}
& \orig && \underline{72.9} & \textbf{\underline{65.7}} &  & 64.9 & 59.1 & \textbf{\underline{88.4}} &  & 86.5 \\
& \ppfi && 56.6 & 57.2 &  & \textbf{\underline{80.6}} & \underline{67.8} & 87.7 &  & \textbf{\underline{86.6}} \\
\midrule
\multirow{3}{*}{\shortstack{\xlmr}}
& \orig && \textbf{\underline{78.4}} & \underline{56.8} &  & 67.3 & 61.2 & \underline{86.8} &  & 85.3 \\
& \ppfi && 60.6 & 51.7 &  & 76.7 & 64.6 & 86.2 &  & \underline{85.4} \\
& \mtfi && 64.3 & 50.3 &  & \underline{77.8} & \textbf{\underline{68.5}} & 86.4 &  & 85.3 \\
\bottomrule
\end{tabular}
}

\newcommand{\insertClassDistributionTable}{
\begin{tabular}[b]{clrcccrccc}
\toprule
&&& \multicolumn{3}{c}{EN} && \multicolumn{3}{c}{EN $\rightarrow$ XX (avg)} \\
\cmidrule{4-6} \cmidrule{8-10}
Model & Train && ent & neu & con && ent & neu & con \\
\midrule
\multirow{2}{*}{\shortstack{\roberta\\\small{({\it{translate-test}})}}}
& \orig && 33.4 & 32.8 & 33.8 &  & 23.2 & 40.7 & 36.1\\
& \ppfi && 34.5 & 31.9 & 33.6 &  & 30.2 & 35.7 & 34.1\\
\midrule
\multirow{3}{*}{\shortstack{\xlmr\\\small{({\it{zero-shot}})}}}
& \orig && 32.4 & 33.2 & 34.4 &  & 27.0 & 37.8 & 35.2\\
& \ppfi && 34.3 & 31.6 & 34.1 &  & 33.1 & 32.9 & 34.0\\
& \mtfi && 33.6 & 32.6 & 33.9 &  & 30.8 & 35.3 & 33.9\\
\midrule
\multicolumn{2}{c}{Gold Standard} && 33.3 & 33.3 & 33.3 && 33.3 & 33.3 & 33.3 \\
\bottomrule
\end{tabular}
}

\newcommand{\insertBiasTable}{
\begin{tabular}[b]{clccc}
\toprule
Model & Train & Base & Unbias & $+\Delta$ \\
\midrule
\multirow{2}{*}{\shortstack{\roberta\\\small{({\it{translate-test}})}}}
& \orig & 77.7\scriptsize{ \textpm 0.6} & 80.6\scriptsize{ \textpm 0.2} & 2.9\scriptsize{ \textpm 0.5} \\
& \ppfi & 82.3\scriptsize{ \textpm 0.2} & 82.8\scriptsize{ \textpm 0.1} & 0.4\scriptsize{ \textpm 0.2} \\
\midrule
\multirow{3}{*}{\shortstack{\xlmr\\\small{({\it{zero-shot}})}}}
& \orig & 81.0\scriptsize{ \textpm 0.2} & 82.4\scriptsize{ \textpm 0.2} & 1.4\scriptsize{ \textpm 0.3} \\
& \ppfi & 83.2\scriptsize{ \textpm 0.1} & 83.3\scriptsize{ \textpm 0.1} & 0.1\scriptsize{ \textpm 0.1} \\
& \mtfi & 83.4\scriptsize{ \textpm 0.2} & 83.8\scriptsize{ \textpm 0.1} & 0.4\scriptsize{ \textpm 0.2} \\
\bottomrule
\end{tabular}
}

\newcommand{\insertXnliSotaTable}{
\begin{tabular}[b]{l|ccccccccccccccc|c}
\toprule
Model & en & fr & es & de & el & bg & ru & tr & ar & vi & th & zh & hi & sw & ur & avg \\
\midrule
\multicolumn{17}{l}{\it Fine-tune an English model and machine translate the test set into English (TRANSLATE-TEST)} \\
\midrule
BERT \citep{devlin2019bert} & 88.8 & 81.4 & 82.3 & 80.1 & 80.3 & 80.9 & 76.2 & 76.0 & 75.4 & 72.0 & 71.9 & 75.6 & 70.0 & 65.8 & 65.8 & 76.2 \\
Roberta \citep{liu2019roberta} & \textbf{\underline{91.3}} & 82.9 & 84.3 & 81.2 & 81.7 & 83.1 & 78.3 & 76.8 & 76.6 & 74.2 & 74.1 & 77.5 & 70.9 & 66.7 & 66.8 & 77.8 \\
Proposed (\roberta\ -- \ppfi) & 90.6 & 85.4 & 86.3 & 84.3 & 85.2 & 85.7 & 82.3 & 80.6 & 81.5 & 77.8 & 78.6 & 81.2 & 77.1 & 73.5 & 72.3 & 81.5 \\
\quad + Unbiasing (tuned in dev) & 90.5 & \textbf{\underline{85.8}} & \textbf{\underline{86.6}} & \underline{84.6} & \textbf{\underline{85.5}} & \underline{85.8} & \underline{82.9} & \underline{81.2} & \underline{82.3} & \underline{78.7} & \underline{79.7} & \underline{82.3} & \underline{77.6} & \underline{74.4} & \underline{72.9} & \underline{82.1} \\
\midrule
\multicolumn{17}{l}{\it Fine-tune a multilingual model on all machine translated training sets (TRANSLATE-TRAIN-ALL)} \\
\midrule
Unicoder \citep{huang2019unicoder} & 85.6 & 81.1 & 82.3 & 80.9 & 79.5 & 81.4 & 79.7 & 76.8 & 78.2 & 77.9 & 77.1 & 80.5 & 73.4 & 73.8 & 69.6 & 78.5 \\
XLM-R \citep{conneau2019xlmr} & \underline{88.7} & \underline{85.2} & \underline{85.6} & \underline{84.6} & \underline{83.6} & \underline{85.5} & \underline{82.4} & \underline{81.6} & \underline{80.9} & \textbf{\underline{83.4}} & \textbf{\underline{80.9}} & \textbf{\underline{83.3}} & \underline{79.8} & \underline{75.9} & \underline{74.3} & \underline{82.4} \\
\midrule
\multicolumn{17}{l}{\it Fine-tune a multilingual model on the English training set (ZERO-SHOT)} \\
\midrule
mBERT \citep{devlin2019bert} & 82.1 & 73.8 & 74.3 & 71.1 & 66.4 & 68.9 & 69.0 & 61.6 & 64.9 & 69.5 & 55.8 & 69.3 & 60.0 & 50.4 & 58.0 & 66.3  \\
XLM \citep{conneau2019xlm} & 85.0 & 78.7 & 78.9 & 77.8 & 76.6 & 77.4 & 75.3 & 72.5 & 73.1 & 76.1 & 73.2 & 76.5 & 69.6 & 68.4 & 67.3 & 75.1 \\
Unicoder \citep{huang2019unicoder} & 85.1 & 79.0 & 79.4 & 77.8 & 77.2 & 77.2 & 76.3 & 72.8 & 73.5 & 76.4 & 73.6 & 76.2 & 69.4 & 69.7 & 66.7 & 75.4 \\
XLM-R \citep{conneau2019xlmr} & \underline{88.8} & 83.6 & 84.2 & 82.7 & 82.3 & 83.1 & 80.1 & 79.0 & 78.8 & 79.7 & 78.6 & 80.2 & 75.8 & 72.0 & 71.7 & 80.1 \\
Proposed (\xlmr\ -- \mtfi) & \underline{88.8} & 84.8 & 85.7 & 84.6 & 84.2 & 85.7 & 82.9 & 81.8 & 82.0 & 82.1 & 79.9 & 81.8 & 79.8 & 75.9 & 76.7 & 82.4 \\
\quad + Unbiasing (tuned in dev) & 88.7 & \underline{85.0} & \underline{86.1} & \textbf{\underline{84.8}} & \underline{84.8} & \textbf{\underline{86.1}} & \textbf{\underline{83.5}} & \textbf{\underline{82.2}} & \textbf{\underline{82.4}} & \underline{83.0} & \underline{80.8} & \underline{82.6} & \textbf{\underline{80.3}} & \textbf{\underline{76.0}} & \textbf{\underline{77.3}} & \textbf{\underline{82.9}} \\
\bottomrule
\end{tabular}
}

%% file: tables_qa.tex
\newcommand{\insertMlqaTable}{
\begin{tabular}[b]{cl|ccccccc|c}
\toprule
Model & Train & en & es & de & ar & vi & zh & hi & avg \\
\midrule
\multicolumn{2}{c}{} & \multicolumn{7}{c}{\it Test set machine translated into English (TRANSLATE-TEST)} & \\
\midrule
\multirow{3}{*}{\shortstack{\roberta}}
& \orig & \textbf{\underline{84.7}} / \textbf{\underline{71.4}} & 70.1 / 49.7 & 60.5 / 41.2 & 55.7 / 32.5 & 65.6 / 40.8 & 53.5 / 26.0 & 42.7 / 20.7 & 61.8{\scriptsize{ \textpm 0.1}} / 40.3{\scriptsize{ \textpm 0.2}}\\
& \ppes & 84.4 / 71.2 & \underline{70.9} / \underline{50.7} & 61.0 / 41.6 & \underline{56.5} / 33.3 & 66.7 / 41.8 & 54.4 / 27.1 & \underline{43.0} / 21.1 & \underline{62.4}{\scriptsize{ \textpm 0.1}} / \underline{41.0}{\scriptsize{ \textpm 0.2}}\\
& \ppfi & 83.8 / 70.4 & 70.3 / 50.1 & \underline{61.1} / \underline{41.9} & \underline{56.5} / \underline{33.4} & \underline{66.8} / \underline{42.1} & \underline{54.9} / \underline{27.5} & 42.8 / \underline{21.3} & 62.3{\scriptsize{ \textpm 0.1}} / 40.9{\scriptsize{ \textpm 0.2}}\\
\midrule
\multirow{5}{*}{\shortstack{\xlmr}}
& \orig & \underline{84.1} / \underline{71.0} & 69.9 / 49.2 & 60.8 / 42.5 & 55.2 / 31.8 & 65.4 / 40.6 & 54.3 / 27.8 & 43.6 / 21.3 & 61.9{\scriptsize{ \textpm 0.1}} / 40.6{\scriptsize{ \textpm 0.1}}\\
& \ppes & 83.8 / 70.8 & \underline{70.5} / \underline{50.0} & \underline{61.4} / \underline{43.5} & \underline{56.1} / \underline{33.1} & \underline{66.5} / \underline{41.6} & 55.4 / 29.0 & \underline{44.0} / \underline{22.2} & \underline{62.5}{\scriptsize{ \textpm 0.2}} / \underline{41.5}{\scriptsize{ \textpm 0.2}}\\
& \ppfi & 82.7 / 69.6 & 70.0 / 49.7 & 61.1 / 43.3 & 56.0 / \underline{33.1} & 66.2 / 41.5 & \underline{55.6} / \underline{29.2} & 43.7 / 22.0 & 62.2{\scriptsize{ \textpm 0.1}} / 41.2{\scriptsize{ \textpm 0.2}}\\
& \mtes & 83.4 / 69.7 & 70.0 / 49.1 & 61.0 / 42.7 & 55.6 / 32.2 & 65.9 / 40.9 & 54.9 / 28.1 & 43.9 / 21.6 & 62.1{\scriptsize{ \textpm 0.3}} / 40.6{\scriptsize{ \textpm 0.2}}\\
& \mtfi & 82.6 / 69.0 & 69.7 / 48.6 & 61.0 / 42.8 & 55.7 / 32.3 & 65.8 / 40.9 & 54.8 / 27.9 & 43.9 / 21.6 & 61.9{\scriptsize{ \textpm 0.3}} / 40.4{\scriptsize{ \textpm 0.2}}\\
\midrule
\multicolumn{2}{c}{} & \multicolumn{7}{c}{\it Test set in target language (ZERO-SHOT)} & \\
\midrule
\multirow{5}{*}{\shortstack{\xlmr}}
& \orig & \underline{84.1} / \underline{71.0} & 74.5 / 56.3 & 70.3 / 55.1 & 66.5 / 45.9 & 74.3 / 53.1 & 67.8 / 43.4 & 71.6 / 53.4 & 72.7{\scriptsize{ \textpm 0.1}} / 54.0{\scriptsize{ \textpm 0.1}}\\
& \ppes & 83.8 / 70.8 & 74.7 / 56.8 & 70.3 / \textbf{\underline{55.2}} & 66.9 / \textbf{\underline{46.5}} & 74.3 / 53.0 & \textbf{\underline{68.2}} / \textbf{\underline{43.8}} & 71.4 / \textbf{\underline{53.6}} & 72.8{\scriptsize{ \textpm 0.2}} / \textbf{\underline{54.3}}{\scriptsize{ \textpm 0.2}}\\
& \ppfi & 82.7 / 69.6 & 74.1 / 56.3 & 69.8 / 54.5 & 66.6 / 46.0 & 73.3 / 52.3 & 67.9 / 43.4 & 71.0 / 53.2 & 72.2{\scriptsize{ \textpm 0.2}} / 53.6{\scriptsize{ \textpm 0.2}}\\
& \mtes & 83.4 / 69.7 & \textbf{\underline{75.2}} / \textbf{\underline{57.3}} & \textbf{\underline{70.5}} / 55.1 & \textbf{\underline{67.5}} / \textbf{\underline{46.5}} & \textbf{\underline{74.5}} / \textbf{\underline{53.2}} & 67.5 / 42.5 & 71.7 / 52.7 & \textbf{\underline{72.9}}{\scriptsize{ \textpm 0.3}} / 53.9{\scriptsize{ \textpm 0.4}}\\
& \mtfi & 82.6 / 69.0 & 74.1 / 56.0 & 70.2 / 54.6 & 66.9 / 46.0 & 73.7 / 52.6 & 67.2 / 41.5 & \textbf{\underline{71.9}} / 53.4 & 72.4{\scriptsize{ \textpm 0.2}} / 53.3{\scriptsize{ \textpm 0.4}}\\
\bottomrule
\end{tabular}
}

\newcommand{\insertXquadFTable}{
\begin{tabular}[b]{cl|ccccccccccc|c}
\toprule
Model & Train & en & es & de & el & ru & tr & ar & vi & th & zh & hi & avg \\
\midrule
\multirow{5}{*}{\shortstack{\xlmr\\\small{({\it{zero-shot}})}}}
& \orig & \textbf{\underline{88.2}} & 82.7 & \textbf{\underline{80.8}} & 80.9 & 80.1 & 76.1 & 76.0 & 80.1 & 75.4 & 71.9 & 76.4 & 79.0{\scriptsize{ \textpm 0.2}}\\
& \ppes & 87.9 & 83.5 & 80.5 & \textbf{\underline{81.2}} & \textbf{\underline{80.7}} & \textbf{\underline{76.8}} & \textbf{\underline{77.4}} & 80.2 & 76.4 & 73.0 & 76.9 & \textbf{\underline{79.5}}{\scriptsize{ \textpm 0.3}}\\
& \ppfi & 87.1 & 82.5 & 80.2 & 80.7 & 79.8 & 75.7 & 76.6 & 79.4 & 75.7 & 71.5 & 76.8 & 78.7{\scriptsize{ \textpm 0.3}}\\
& \mtes & 87.1 & \textbf{\underline{84.1}} & 80.3 & \textbf{\underline{81.2}} & 80.1 & 76.0 & \textbf{\underline{77.4}} & \textbf{\underline{80.9}} & 76.7 & 72.7 & 77.1 & 79.4{\scriptsize{ \textpm 0.3}}\\
& \mtfi & 86.3 & 81.4 & 80.2 & 80.5 & 80.2 & 76.6 & 77.0 & 80.3 & \textbf{\underline{77.6}} & \textbf{\underline{74.5}} & \textbf{\underline{77.8}} & 79.3{\scriptsize{ \textpm 0.2}}\\
\bottomrule
\end{tabular}
}